\documentclass{article}
\usepackage[utf8]{inputenc}
\pdfoutput=1
\usepackage{ms}  % DO NOT CHANGE THIS
\usepackage{times}  % DO NOT CHANGE THIS
\usepackage{helvet} % DO NOT CHANGE THIS
\usepackage{courier}  % DO NOT CHANGE THIS
\usepackage[pdftex]{graphicx} % DO NOT CHANGE THIS
\usepackage{algorithm}
\usepackage{algorithmic}
\usepackage{amsmath}
\usepackage{subfigure}	
\usepackage{caption}
\usepackage{natbib}
\title{Hybrid Learning with New Value Function for the Maximum \\ Common Subgraph Problem}
\author{Yanli Liu,\textsuperscript{\rm 1}
Jiming Zhao\textsuperscript{\rm 1}
Chu-Min Li,\textsuperscript{\rm 2}\thanks{Corresponding author.}
Hua Jiang,\textsuperscript{\rm 3}
Kun He\textsuperscript{\rm 4}\\
\textsuperscript{\rm 1}WuHan University of Science and Technology, China \\
\textsuperscript{\rm 2}MIS, Universit\'e de Picardie Jules Verne, France\\
\textsuperscript{\rm 3} Yunnan University, China\\
\textsuperscript{\rm 4} Huazhong University of Science and Technology, China \\}

\begin{document}

\maketitle

\begin{abstract}

Maximum Common induced Subgraph (MCS) is an important NP-hard problem with wide real-world applications. Branch-and-Bound (BnB) is the basis of a class of efficient algorithms for MCS, consisting in successively selecting vertices to match and pruning when it is discovered that a solution better than the best solution found so far does not exist. The method of selecting the vertices to match is essential for the performance of BnB. In this paper, we propose a new value function and a hybrid selection strategy used in reinforcement learning to define a new vertex selection method, and propose a new BnB algorithm, called McSplitDAL, for MCS. Extensive experiments show that McSplitDAL significantly improves the current best BnB algorithms, McSplit+LL and McSplit+RL. An empirical analysis is also performed to illustrate why the new value function and the hybrid selection strategy are effective.
\end{abstract}

\section{Introduction}
Graphs have gained increasing attention in recent decades due to their natural expression in representing numerous real-world problems. Given two graphs $G_p (V_p,E_p)$ and $G_t (V_t,E_t)$,
the {\em Maximum Common induced Subgraph} (MCS) problem is to find a subgraph $G_p'$ of $G_p$ and a subgraph $G_t'$ of $G_t$, such that $G_p'$ and $G_t'$ are isomorphic and have the maximum number of vertices. MCS allows to evaluate the similarity of two graphs and has broad applications in many domains, such as graph database systems~\cite{c:database05}, biochemistry~\cite{ar:bio2013,ar:bio2017}, malware detection~\cite{ar:Park2013,ar:Ad2021}, cheminformatics~\cite{ar:John2002,ar:Ramon2020,ar:Robert2021}, computer vision~\cite{ar:Solnon2015}, communication networks~\cite{ar:Paris2016}, \textit{etc}. There are also many MCS variant problems, such as the Maximum Common Connected induced Subgraph (MCCS) problem and the Subgraph Isomorphic (SI) problem. %\textit{etc}.
%image or video analysis~\cite{c:Bunke1995,Liu2001Graph}, and pattern recognition~ \cite{D2008THIRTY,Cordella2004A}. image analysis~\cite{ar:Liu2001Graph},

MCS is NP-hard and thus computationally challenging.
Despite its NP-hardness, many methods have been developed to solve MCS, including exact and inexact algorithms. An important class of exact methods exploits the powerful branch-and-bound (BnB) framework~\cite{ar:John2002, c:McCreesh2017, c:LIU2020, c:zhou2022} to travel the whole search tree
and try to match each vertex of $G_p$ with each vertex of $G_t$ (\textit{i.e.}, branching) to find the best matches.
The key to design an efficient BnB algorithm is to reduce the search space, using techniques such as effective branching heuristic~\cite{ar:Peter2015, c:zhou2022} and powerful constraint filtering~\cite{ar:Solnon2015, c:McCreesh2016, c:McCreesh2017, ar:Robert2021}. When exact solutions are not required, one can use inexact algorithms to find approximate solutions of acceptable quality without exhausting the search space. These inexact methods include meta-heuristics~\cite{c:Rutgers2010, c:Choi2012} and spectra methods~\cite{ar:Edwin2009}. Recently, classification technologies of machine learning~\cite{c:Andrei2018, c:Li2020} and graph neural networks~\cite{c:Bai2021} are also adopted to solve MCS.

Some state-of-the-art BnB algorithms combine the advantages of both search and reinforcement learning techniques to improve their branching methods to efficiently reduce the search space. Based on the BnB algorithm framework, McSplit~\cite{c:McCreesh2017} uses a partition method to satisfy the isomorphic constraint and a branching heuristic based on the obtained partition and vertex degree to minimize the search tree size. McSplit+RL~\cite{c:LIU2020} explores the vertex pair selection policy based on reinforcement learning with a value function for each vertex so that it can reach a leaf node of search tree as early as possible. McSplit+LL~\cite{c:zhou2022} further proposes a long-short memory and leaf vertex union match to improve the performance of a BnB MCS algorithm.

We observe that the learning policies in McSplit+RL and McSplit+LL only concentrate on the reduction of upper bound due to a branching. They only reward branching vertices of input graphs with upper bound reduction and select a new branching vertices in the decreasing order of their accumulative rewards, that can make the algorithm to branch only on a small set of vertices so as to get trapped around local optima.

To remedy these limitations, we propose a new value function, namely Domain Action Learning (DAL), for evaluating each branching, that considers both upper bound reduction and real graph simplification due to a branching action.
We further propose a hybrid vertex selection strategy based on different value functions to guide the search. In fact, the best branching selected by different value functions is usually different. Branching alternatively according to two value functions allows to effectively diversify the search.

Based on the above new value function and the hybrid vertex selection strategy, we propose a new BnB algorithm, termed McSplitDAL, on top of McSplit~\cite{c:McCreesh2017}. Experiments are conducted to evaluate McSplitDAL on 24,761 instances derived from diverse applications.
The experimental results show that McSplitDAL significantly outperforms McSplit+RL and McSplit+LL that are already highly efficient. We also conduct empirical analysis and provide insight on why the proposed algorithm is effective.

This paper is organized as follows. Section 2 gives some basic graph definitions and related concepts used in this paper. Section 3 reviews existing BnB algorithms and learning methods for MCS. Section 4 describes our new value function and hybrid vertex selection policy, followed by our McSplitDAL algorithm. Section 5 presents the empirical results and analysis. Section 6 concludes.

%The main contributions of this paper are summarized below.
%The rest of the paper is organized as follows
%as the baselines of McSplit+RL and McSplit+LL.
%We exclude the easy instances that can be solved by all the three algorithms (McSplit+RL, McSplit+LL, and McSplitDAL) within 10 seconds, and the too hard instances that none of the three algorithms can solve within the cut-off time 1800 seconds. In the end, there are 2,229 remained medium instances. The results show that McSplitDAL solves 16.3$\%$ instances more than McSplit+LL and $28.6\%$ instances more than McSplit+RL,

\section{Preliminaries}
Consider a simple, undirected and unlabelled graph $G = (V,E)$, where $V$ is the set of vertices, $E\subseteq V\times V$ is the set of edges. Two vertices $u$ and $v$ are adjacent (or neighbors) if $(u, v) \in E$. The degree of a vertex $v$ is the number of its adjacent vertices. A subgraph of $G$ induced by a vertex subset $V' \subseteq V$ is defined by $G[V'] = (V',E')$,  where $E'=\{(u, v) \in E | u, v \in V'\}$.
%When it is clear from the context, we also use $V'$ to denote the subgraph induced by $V'$.

Given a graph $G_p = (V_p, E_p)$ (named pattern graph) and a graph $G_t = (V_t, E_t)$ (named target graph), if there exist an induced subgraph $G'_p=(V'_p, E'_p)$ of $G_p$, an induced subgraph $G'_t=(V'_t, E'_t)$ of $G_t$, and a bijection $\phi: V'_p \rightarrow V'_t$,
such that any $v$ and $v'$ of $V'_p$ are adjacent in $G_p$ if and only if $\phi(v)$ and $\phi(v')$ of $V'_t$ are adjacent in $G_t$,
then $G'_p$ or $G'_t$ is called an induced common subgraph of $G_p$ and $G_t$. In this case, we say that $v$ and $\phi(v)$ are matched and $(v, \phi(v))$ is a match.  The {\em Maximum Common induced Subgraph (MCS) problem} is to find a common induced subgraph of $G_p$ and $G_t$ with the maximum number of vertices.
Let $V'_p=\{v_1, v_2, \ldots, v_{|V'_p|}\}$, a feasible solution of MCS can be represented as a set of matched pairs $\{(v_1, w_1), (v_2,w_2), \ldots, (v_{|V'_p|}, w_{|V'_p|})\}$, where $w_j = \phi(v_j))$ for $j \in \{1, ..., {|V'_p|}\}$.

A variant of MCS called the Maximum Common Connected induced Subgraph (MCCS) problem requires that the maximum common induced subgraph is connected. Another variant called Subgraph Isomorphism (SI) requires $V'_p = V_p$.

A common induced subgraph $G'_p$ is maximal if it cannot be extended to a larger common induced graph. If a feasible solution $\{(v_1, w_1), (v_2,w_2), \ldots, (v_{|V'_p|}, w_{|V'_p|})\}$ is not maximal, it induces a nonempty set of $s$ vertex subset pairs
$Ev = \{(V_{1p}, V_{1t}), \ldots, (V_{sp}, V_{st})\}$, where for $1\leq i \leq s$, $V_{ip}$ ($V_{it}$) is a subset of $V_p$ ($V_t$) and for any $i'\neq i$, $V_{i'p}$ ($V_{i't}$) is disjoint with $V_{ip}$ ($V_{it}$), with the following property~\cite{c:McCreesh2017}:

\begin{itemize}
\item
For any $1\leq i \leq s$ and any $1\leq j \leq {|V'_p|}$, either all vertices in $V_{ip}$ are adjacent to $v_j$ and all vertices in $V_{it}$ are adjacent to $w_j$, or all vertices in $V_{ip}$ are non-adjacent to $v_j$ and all vertices in $V_{it}$ are non-adjacent to $w_j$.
\end{itemize}

\begin{figure}
	\centering
	\includegraphics[width=3.3in]{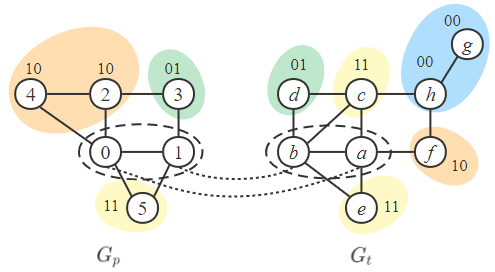}
	\caption {An illustration of an environment of MCS and its related concepts. A non-maximal common induced subgraph $\{(0,a), (1,b)\}$ induces a set of vertex subset pairs (called an environment) $Ev$ = $\{$ ($\langle 3 \rangle$, $\langle d \rangle$), ($\langle 4, 2 \rangle$, $\langle f \rangle$), ($\langle 5 \rangle$, $\langle c, e \rangle$) $\}$, in which the vertex subset pair ($\langle 4, 2 \rangle, \langle f \rangle$) is labelled `10', because vertices 4 and 2 are all adjacent to 0 and $f$ is adjacent to $a$, while 4 and 2 are all non-adjacent to 1 and $f$ is non-adjacent to $b$. Other labels in the graphs are interpreted similarly. Vertices with the same color (or the same label) are in the same domain. Note that there is no vertex labelled with `00` in $G_p$, so that vertices labelled with `00' in $G_t$ cannot be matched with any vertex in $G_p$. In fact,
	this $Ev$ can provide at most 3 additional matches.
	}
\label{Figure1}
\end{figure}

As illustrated in Figure \ref{Figure1},
if $V_{ip}$ and $V_{it}$ are not empty for some $1\leq i \leq s$, choosing any vertex $v$ in $V_{ip}$ and $w$ in $V_{it}$ allows to extend the induced common subgraph by this vertex pair. Clearly, $Ev$ can be defined to be $\{(V_{p}, V_{t})\}$ when the common induced subgraph is empty.

For any $1\leq i\leq s$ and any $1\leq j \leq {|V'_p|}$, since all vertices in $V_{ip}$ and in $V_{it}$ have the same (non-)adjacency to $v_j$ and $w_j$, we can use a bit 1 (0) to say that all vertices in $V_{ip}$ or $V_{it}$ are (non-)adjacent to $v_j$ or $w_j$, respectively. So, a vertex subset pair $(V_{ip}, V_{it})$
can be labelled using a ${|V'_p|}$-bit string, in which the $j^{th}$ bit indicates whether vertices in $V_{ip}$ and $V_{it}$ are adjacent to $v_j$ and $w_j$~\cite{c:McCreesh2017}.

 If for some $1\leq i_1 < i_2 \leq s$, the pairs $(V_{i_1p}, V_{i_1t})$ and $(V_{i_2p}, V_{i_2t})$
 have the same label, then they should be combined into one pair. So, we assume the labels in $Ev$ are distinct. Furthermore, any pair $(V_{ip}, V_{it})$ in which $V_{ip}$ or $V_{it}$ is empty is removed from $Ev$.

 Consequently, for any $1\leq i_1 < i_2 \leq s$, any vertex $v$ in $V_{i_1p}$ ($V_{i_2p}$) cannot be matched with any vertex $w$ in $V_{i_2t}$ ($V_{i_1t}$) such that $G'_p=G_p[\{v_1, \ldots, v_{|V'_p|}\}]$ extended with $v$ and $G'_t=G_t[\{w_1, \ldots, w_{|V'_p|}\}]$ extended with $w$ remain isomorphic, because there is a $j$ ($1\leq j\leq |V'_p|$) such that $v$ is adjacent to $v_j$, but $w$ is not adjacent to $w_j$, or vice versa.
 %So, a labelled pair $(V_{ip}, V_{it})$ can offer at most $min(|V_{ip}|, |V_{it}|)$ additional and legal matches to $D$.
 Thus, the sum $\sum_{(V_{ip}, V_{it}) \in Ev} \min(|V_{ip}|, |V_{it}|)$ provides an upper bound of the number of vertices that can be added into the common induced subgraph $G'_p$ and $G'_t$.

 In this paper, each labelled pair $(V_{ip}, V_{it})$ is called a domain, because it specifies two sets of vertices that can be matched, and $Ev$ is called an environment in which a BnB MCS algorithm works.

\section{Search for MCS and Learning Policy}
This section first presents a state-of-the-art BnB search framework as shown in Algorithm 1, which allows an exploration of search space and enforces the isomorphism constraint. Thus, it serves as the backbone of McSplit+RL \cite{c:LIU2020}, McSplit+LL \cite{c:zhou2022} and our McSplitDAL. Then, we review representative learning policies that tell a BnB algorithm how to select a branching pair.
%From the view of Reinforcement Learning (RL), two input graphs are the RL environment and a search algorithm is an agent that decides how to get an increasing isomorphic subgraph, whose action is to select a vertex pair from the remaining subgraphs.

\subsection{Branch and Bound for MCS}
To simplify the description, we suppose that two input graphs are undirected and unlabelled, and search methods can be easily extended to other kinds of graphs~\cite{c:McCreesh2016}.

Given a pattern graph $G_p$ and a target graph $G_t$, the BnB algorithm MCS depicted in Algorithm \ref{alg:BnB} works with an environment $Ev$ (i.e., a set of domains), a policy $\pi_v$ to select a vertex in $G_p$, a policy $\pi_w$ to select a vertex $w$ in $G_t$ to match with $v$, a current growing solution $curSol$, and the best solution $MaxSol$ found so far. At the beginning, $curSol$ and $MaxSol$ are both empty, and $Ev = \{(V_p, V_t)\}$ contains only one domain, meaning that every vertex in $V_p$ is a candidate to match every vertex in $V_t$.

MCS first estimates an upper bound on the number of matches that can be found with the current $Ev$.
%each vertex in $V_p$ is regarded as a CSP variable associated with a value domain initiated to be $V_t$, meaning that each vertex in $V_p$ can match each vertex in $V_t$. The BnB algorithm first estimates an upper bound of the number of the matching vertex pairs satisfying the isomorphic constraint that can be found in the current subtree.
If the $UB$ is not larger than the size of the best solution $MaxSol$ found so far, the algorithm prunes this branch and backtracks (Line \ref{UB}-- \ref{UBend}). Otherwise, the algorithm selects a new vertex pair $(v,w)$ such that $v\in V_p$ and $w\in V_t$ to match using policy $\pi_v$ and $\pi_w$ respectively. As a consequence of matching $v$ with $w$, $(v,w)$ is added into $curSol$, and each domain in $Ev$ is split into two domains $D_1$ ($D_2$): domain in which each vertex of $V_p$ is (non-)adjacent to $v$ and each vertex of $V_t$ is (non-)adjacent to $w$ (Line \ref{update}--\ref{updateend}). Domains with at least one empty vertex subset are removed.
%After splitting, any new domain $(V'_p, V'_t)$ with empty $V'_p$ or $V'_t$ is removed from $Ev$.
Afterwards, the algorithm runs recursively on the new domains (Line \ref{recur}). After finishing the search of the subtree rooted at $(v,w)$, the algorithm tries to match $v$ with other vertices in $V_t$ (Line \ref{backtrack}) selected using policy $\pi_w$. Then, it removes $v$ from  $Ev$ and runs recursively. At last, the optimal solution is returned (Line \ref{otherway}--\ref{return}).

\begin{algorithm}[tb]
\caption{MCS$(Ev, \pi_v, \pi_w, curSol, MaxSol)$}
\label{alg:BnB}
\textbf{Input}: a domain set $Ev$; policies $\pi_v$ and $\pi_w$ for selecting the matching pair $(v,w)$; the current solution $curSol$ and the best solution found so far $MaxSol$\\
\textbf{Output}: $MaxSol$
\begin{algorithmic}[1] %[1] enables line numbers
\STATE $UB \leftarrow |curSol|$ + $\sum_{(V_{ip}, V_{it}) \in Ev} \min(|V_{ip}|, |V_{it}|)$      \label{UB}
\IF {$UB \leq |MaxSol|$}
    \STATE \textbf{return} $MaxSol$
\ENDIF                                                    \label{UBend}
\STATE $(V_{ip},V_{it}) \leftarrow selectD(Ev)$            \label{selectpaire}
%\STATE $D_p \leftarrow $ all vertices $\in V_p$ in $D$
%\STATE $D_t \leftarrow $ all vertices $\in V_t$ in $D$
\STATE $v \leftarrow$ $selectV(V_{ip}$, $\pi_v$)              \label{select}
\FOR {$k$ in range($|V_{it}|$)}
   \STATE $w \leftarrow selectW(V_{it}, \pi_w)$
   \STATE $V_{it} \leftarrow$ $V_{it}$  $\backslash \{ w \}$   \label{selectend}
   \STATE $curSol \leftarrow$ $curSol \cup \{(v, w)\}$      \label{update}
   \IF {$|curSol| > |MaxSol|$}
       \STATE $MaxSol \leftarrow curSol$
   \ENDIF
   \STATE $Ev' \leftarrow$ a new domain set obtained by splitting domains in $Ev$      \label{updateend}
   \STATE $MaxSol \leftarrow MCS(Ev',\pi_v, \pi_w, curSol, MaxSol)$        \label{recur}
   \STATE $curSol \leftarrow$ $curSol \backslash \{(v, w)\}$                \label{backtrack}
\ENDFOR
\STATE $Ev' \leftarrow$ a new domain set
by removing $v$ from $Ev$ \label{otherway}
\STATE $MaxSol \leftarrow MCS(Ev',\pi_v, \pi_w, curSol, MaxSol)$
\STATE \textbf{return} $MaxSol$                                             \label{return}
\end{algorithmic}
\end{algorithm}

The policies $\pi_v$ and $\pi_w$ are both based on a $selectD(\cdot)$ function that returns a domain from $Ev$, in which $v$ and $w$ are selected.  \cite{c:McCreesh2017} provide a $selectD(\cdot)$ function, by defining the size of a domain ($V_{ip},V_{it}$) to be max($|V_{ip}|,|V_{it}|$) and returning the domain with the smallest size from $Ev$, with ties broken by the largest vertex degree in $V_{ip}$. This function is used in Algorithm \ref{alg:BnB}.

\subsection{Related Learning Policy}
In a BnB algorithm for MCS, the branching heuristic to select $v$ and $w$ to match is crucial to reduce the size of the search tree. Early heuristics mainly rely on the properties of input graphs ~\cite{ar:Solnon2015,ar:Peter2015,ar:varorder2017,c:McCreesh2017} and focus on selecting $v$, while $w$ is selected in turn to be matched with $v$ in their natural order or the decreasing degree order. For instance, the degree heuristic first matches the vertex with the highest degree~\cite{ar:Solnon2015}. The degree-weighted-domains heuristic is to select a vertex with the greatest degree in the smallest domain~\cite{c:BoussemartHLS04}. The neighbourhood heuristic selects a vertex that is a neighbor of the current partial order of matched pairs~\cite{c:neibor14}. McSplit~\cite{c:McCreesh2017} first selects a domain in $Ev$ with the smallest max($|V_{ip}|,|V_{it}|$) value, and then the vertex $v$ ($w$) with the greatest degree in $V_{ip}$ ($V_{it}$).

%There is not reward to match action or divide vertex.
Recent heuristics use reinforcement learning to improve the branching heuristic of McSplit. They regard the BnB algorithm as an agent having a goal of reaching a search tree leaf as soon as possible. An action of the agent is to match a vertex $v$ in $V_p$ with a vertex $w$ in $V_t$. %However, there are usually many choices of $v$ and $w$ at each step, and the agent does not know in advance which choice is better. So, a value function needs to be defined,
A value function is defined
based on a reward given to each performed action, then reinforcement learning is used to recognize the best action to choose at each step based on the accumulative rewards of each action received in the past. So, the key issue here is how to define a reward and a value function, and how to exploit them to select an action.

 McSplit+RL~\cite{c:LIU2020} defines the reward of matching $(v, w)$ to be the upper bound reduction produced by the matching. Then, both $v$ and $w$ receive this reward. The policy $\pi_v$ selects $v$ with the highest accumulative rewards in the smallest domain (\textit{i.e.}, a domain with the smallest size as defined in McSplit) and $w$ is selected in the same domain in the decreasing order of their accumulative rewards, to be matched with $v$ in turn.

 As can be seen in Algorithm \ref{alg:BnB}, the algorithm reaches a leaf when $UB$ $\leq |MaxSol|$. Therefore, picking a vertex with the greatest accumulated reductions of upper bound can help McSplit+RL reach a leaf quickly. We refer the policies of McSplit+RL to select $v$ and $w$ to match by {\em RL}.

McSplit+LL ~\cite{c:zhou2022} further reduces the size of the search tree with Long-Short Memory (LSM) and Leaf vertex Union Match (LUM) techniques, which employs the same matching reward as McSplit+RL. But McSplit+LL manages the vertex value differently from McSplit+RL. Specifically, LSM records the accumulative rewards of each vertex in $G_p$, and the accumulative rewards of each vertex pair $(v_i,w_j)$ matched in the past. At each step, it picks the vertex $v$ in $G_p$ with the greatest accumulative rewards in the smallest domain, as McSplit+RL does, then picks a vertex $w$ in $G_t$ in the same domain such that the vertex pair ($v$, $w$) has the greatest accumulative rewards among $\{(v,w_1,), (v,w_2), \dots ,(v,w_{|V_t|})$. LUM is to simultaneously match the leaf neighbors of $v$ to the leaf neighbors of $w$ after matching $(v,w)$. The leaf neighbor of a vertex is its neighbor with degree 1.

We observe that there are two limitations in the above learning policies. First, the reward for a matching action is only defined by its effect on upper bound. However, consider two possible matchs $(v, w)$ and $(v', w')$. The graph simplification due to these two matches can be very different, even if they give the same upper bound reduction. Second, these policies tend to produce a kind of ``Matthew effect": the vertices with high accumulated rewards will be chosen again and again upon backtracking, and get their accumulated rewards higher, % and higher,
while the vertices with low %or 0
accumulated rewards have little chance to be chosen and their accumulated rewards stay low. The Matthew effect can make the algorithm mainly branch on a small subset of vertices, so that the search get trapped around local optima.

%To this end, we propose a new value function based on a new reward definition, and a new hybrid vertex selection strategy to overcome these two limitations.
In the next section, we will propose a new value function based on a new reward definition, and a new hybrid vertex selection strategy to overcome these two limitations.

\section{Proposed Method}
In this section, we define a new reward to an action of matching a vertex $v$ in $G_p$ and a vertex $w$ in $G_t$, in order to reflect more accurately the consequence of the action, further obtain a new vertex selection policy. We then propose a hybrid strategy combining the new vertex selection policy with {\em RL}, the policy of McSplit+RL, allowing to overcome the Matthew effect of a single policy.

Note that the tabular method in reinforcement learning is not directly applicable to BnB MCS algorithms, because there are too many states and actions to store, and approximate functions needs lots of computation to fit state value or action value for MCS.

\subsection{New Value Function}

\begin{figure}[t!]
\centering	
\begin{subfigure}
    \centering
    \includegraphics[width=3.2in]{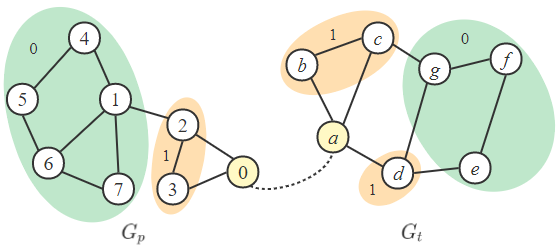}
    \caption*{ (I) $Ev=\{(\langle 2,3 \rangle, \langle b,c,d \rangle), (\langle 1,4,5,6,7 \rangle, \langle e,g,f \rangle)\}$, induced by $\{(0,a)\}$ }
\end{subfigure}
\vspace{1em}
\begin{subfigure}
    \centering
    \includegraphics[width=3.2in]{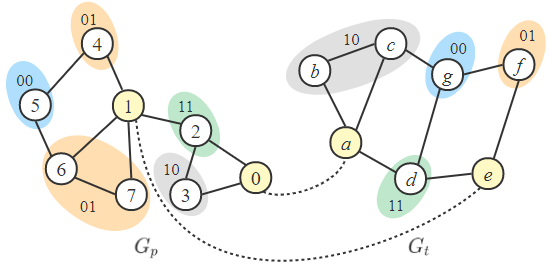}
    \caption*{ (II) $Ev'=\{(\langle 2 \rangle, \langle d \rangle), (\langle 3 \rangle, \langle b,c \rangle), (\langle 4,6,7 \rangle,  \langle f \rangle), (\langle 5 \rangle,$ \\
    $\langle g \rangle)\}$, induced by $\{(0,a),(1,e)\}$ }
\end{subfigure}
\vspace{1em}
\begin{subfigure}
    \centering
    \includegraphics[width=3.2in]{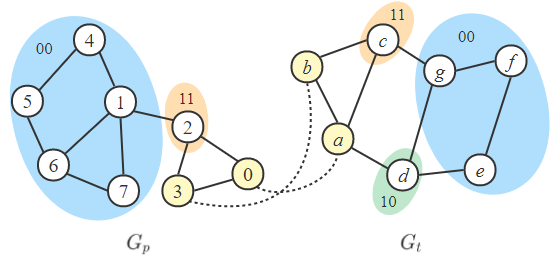}
    \caption*{ (III) $Ev''$ = $\{(\langle 2 \rangle, \langle c \rangle), (\langle 1,4,5,6,7 \rangle, \langle e,g,f \rangle)\}$, induced by $\{(0,a),(3,b)\}$ }
\end{subfigure}
\caption{An example of modifying environment by a match, where $Ev$ = $\{(\langle 2,3 \rangle, \langle b,c,d \rangle), (\langle 1,4,5,6,7 \rangle, \langle e,g,f \rangle)\}$ is induced by matching vertex $0$ with vertex $a$. Then it transforms into a new domain set containing simpler subgraphs by matching vertex 1 with vertex $e$ than by matching vertex 3 with vertex $b$. The vertices with the same color are in the same domain.}
\label{Figure6}
\end{figure}

A BnB algrithm as specified in Algorithm \ref{alg:BnB} works with a set $Ev = \{(V_{1p}, V_{1t}), \ldots, (V_{sp}, V_{st}) \}$ of domains, where each vertex subset pair $(V_{ip}, V_{it})$ ($1\leq i \leq s$) is a domain. We say that $Ev$ is the environment that the learning agent stays. Each environment induces an upper bound of the number of matches that can be added into the current growing solution. An action matching a vertex $v$ and a vertex $w$ changes the current environment. The changed environment induces a new upper bound. The difference between the old upper bound and the new one is used as the reward to the action $(v, w)$ in McSplitRL and McSplitLL. We argue that this reward is not accurate enough for an action, because two actions inducing the same upper bound reduction can result in different domains. Figure \ref{Figure6} gives an illustrative example.

\newtheorem{exam}{Example}
\begin{exam}
Figure \ref{Figure6} shows two graphs $G_p$ and $G_t$, $V_p$ = $\{0,1,2,3,4,5,6,7\}$, $V_t$ = $\{a,b,c,d,e,f,g\}$. The vertex pair (0,a) has been first matched, so that the current $Ev$ = $\{(\langle 2,3 \rangle, \langle b,c,d \rangle), (\langle 1,4,5,6,7 \rangle, \langle e,g,f \rangle)\}$, that can at most provide 2 + 3 = 5 vertex pairs to extend current solution $\{(0,a)\}$ . Thus, the current upper bound is 5.

If the second matching is $(1,e)$, $Ev$ will be modified into $Ev'$ = $\{(\langle 2 \rangle, \langle d \rangle), (\langle 3 \rangle, \langle b,c \rangle), (\langle 4,6,7 \rangle,  \langle f \rangle), (\langle 5 \rangle, \langle g \rangle)\}$. The upper bound induced by $Ev'$ is 4 and $|Ev'|$ = 4.

If the branching strategy picks $(3,b)$ instead of $(1,e)$ as the second matching, $Ev$ will be modified into $Ev''$ = $\{(\langle 2 \rangle, \langle c \rangle), (\langle 1,4,5,6,7 \rangle, \langle e,g,f \rangle)\}$. The upper bound induced by $Ev''$ is 4 and $|Ev''|$ = 2.

The two matches $(1, e)$ and $(3, b)$ induce the same upper bound reduction. However, the environment change from $Ev$ to $Ev'$ induced by the matching $(1, e)$ is clearly more important than the change from $Ev$ to $Ev''$ induced by the matching $(3, b)$, because the domains in $Ev$ are split into more new domains, meaning that the problem is more simplified by the branching on $(1, e)$. We use the number of domains in $Ev'$ or $Ev''$ to measure the environment change. Intuitively, the more the environment changes, the more the subproblem is easier to solve.

%Note that the fact that an environment resulted from the same original environment contains more domains usually means that the contained domains are  smaller.
\label{statereward}
\end{exam}

Based on the above observation, we propose a new reward defined in Equation \ref{reward}, where $Ev'$ is the environment modified from $Ev$ by the match $(v, w)$.

\begin{equation}
\begin{aligned}
R(v,w) = & \sum_{\small{(V_{ip}, V_{it})} \in Ev} \min(|V_{ip}|, |V_{it}|)    -  \\&
           \sum_{\small{(V_{ip}', V_{it}')} \in Ev'} \min(|V_{ip}'|, |V_{it}'|) + |Ev'|
\label{reward}
\end{aligned}
\end{equation}

Equation \ref{reward} uses both the upper bound reduction and the number of domains contained in the new environment $Ev'$ to reward an action $(v, w)$. A new value function called Domain and Action Learning (DAL) is defined by Equation \ref{DALV} and Equation \ref{DALW}.

\begin{equation}
DAL(v) \leftarrow DAL(v) + R(v, w)
\label{DALV}
\end{equation}

\begin{equation}
DAL(v, w) \leftarrow DAL(v, w) + R(v, w)
\label{DALW}
\end{equation}

%Equation \ref{value} based on this reward to evaluate each matched vertex in long runs.

%\begin{equation}
%q(v, w) \leftarrow q(v, w) + R(v, w)
%\label{value}
%\end{equation}

 The DAL value function considers both the upper bound reduction and the number of domains contained in the new environment $Ev'$. A greater upper bound reduction presumably allows to prune the search earlier. A greater number of domains in $Ev'$ presumably implies a subproblem easier to solve. Our purpose is to combine the two advantages to speed up the search.

 A new vertex selection policy is thus defined using the DAL value function, which gives the score $DAL(v)$ for each vertex $v$ in $V_p$ and the score $DAL(v, w)$ for each match $(v, w)$, all initialized to 0. Then at each step, after selecting the smallest domain $(V_{ip},V_{it})$ from the current environment $Ev$ in Algorithm \ref{alg:BnB} (Line \ref{selectpaire}), the vertex $v$ in $V_{ip}$ with the highest $DAL(v)$ is selected, and matched in turn with $w$ in $V_{it}$ in the decreasing order of $DAL(v, w)$. After matching $v$ with each $w$, $R(v, w)$ defined in Equation \ref{reward} is added into $DAL(v)$ and $DAL(v, w)$.

As in McSplitLL, if $DAL(v)$ and $DAL(v, w)$ reach $T_v$ and $T_{vw}$, respectively, where $T_v$ and $T_{vw}$ are two parameters as in McSplitLL, all vertex values in $DAL(v)$ and $DAL(v, w)$ decay to a half.
 %are divided by 2 , when their value .

\subsection{Hybrid Branching Policy}
As is explained in the previous section, the current vertex selection policies based on reinforcement learning can suffer from the Matthew effect, so does the new vertex selection policy based on the {\em DAL} value function.

In order to overcome the Matthew effect, we propose to hybrid the {\em RL} policy of McSplitRL and the {\em DAL} policy defined in this paper. Concretely, let $\Pi \in \{RL, DAL\}$ denote the current policy, and be initialized to be $RL$. Every time $v$ or $w$ is selected using $\Pi$, a counter $NbApp$ is incremented by 1. When $NbApp$ reaches a fixed threshold $MaxNbApp$, it is reset to 0, and $\Pi$ is changed to another policy in $\{RL, DAL\}$. An exception happens when a better solution is found. In this case, $NbApp$ is reset to 0, and the same policy continues to be used.

Note that the vertices with the highest value {\em DAL} or {\em RL} are usually different. The hybrid branching policy allows to branch on different vertices, thus diversifying the search while keeping good quality branchings.

\section{Experiments}
 The proposed algorithm McSplitDAL is implemented in C++ on top of McSplit and compiled using g++ -O3. We conduct experiments to evaluate the new algorithm and the proposed strategies. All experiments were performed on Intel Xeon CPUs E5-2680 v4@2.40 G under Linux with 4G memory.

The three parameters $T_v$, $T_{vw}$ and $MaxNbApp$ are set to $10^5$, $10^9$, $2 \times min(|V_p|,|V_t|)$, respectively.

\subsection{Benchmarks}
The benchmark datasets include 24,761 instances, which are divided into two sets.

$\bullet$ Biochemical reactions ~\cite{ar:Gay2014}: including 136 directed unlabelled bipartite graphs. The number of vertices varies from 9 to 386. All graphs describe the biochemical reaction networks. This dataset provides 9316 instances by pairing any two graphs (including 136 self-match pairs).

$\bullet$ Large SI instances~\cite{ar:ima2011,ar:Solnon2015,c:Huffmann17,c:McCreesh2017,c:LIU2020,c:zhou2022}: including 15,445 instances generated from the real-world problems or random models, such as
segmented images, modelling 3D objects, and scale-free networks. Specifically, this instance set contains: 6,278 {\em Images-CVIU11}, 1225 {\em LV}, 3,430 {\em LargerLV}, 24 {\em Image-PR15}, 1170 {\em SI}, 100 {\em Scalefree}, 3018 {\em Meshes-CVIU11} and 200 {\em phase}. The number of vertices varies from 22 to 6,671.

The time limit  for each instance in the experiments is 1800 seconds.

\subsection{Solvers}

 We compare the new algorithm McSplitDAL with two state-of-the-art BnB algorithms: McSplit+LL ~\cite{c:zhou2022} and McSplit+RL ~\cite{c:LIU2020}. To better understand the proposed polices, four variants of these algorithms are also included in the experiments.

 $\bullet$ McSplitDAL: our implementation of Algorithm \ref{alg:BnB} on top of McSplit ~\cite{c:McCreesh2017} with the new value function {\em DAL} and the hybrid vertex selection policy $\Pi \in \{RL, DAL\}$.

$\bullet$ McSplit+RL ~\cite{c:LIU2020}: An implementation of the Algorithm \ref{alg:BnB} on top of McSplit with the value function {\em RL}, which significantly improves McSplit.

$\bullet$ McSplit+LL ~\cite{c:zhou2022}: An implementation of the Algorithm \ref{alg:BnB} on top of McSplit with the LSM and LUM techniques.

$\bullet$ McSplitRLD: a variant of McSplitRL using the new {\em DAL} value function instead of the {\em RL} policy.

$\bullet$ McSplitLLD: a variant of McSplit+LL using the new {\em DAL} value function instead of its own policy.

$\bullet$ McSplitDAL+rand: A variant of McSplitDAL, which applies one of two branching policies $\{RL, DAL\}$ in random at each branch node, instead of applying each policy $MaxNbApp$ times alternatively.

$\bullet$ McSplitDAL+depth: A variant of McSplitDAL, which changes the policy according to the depth of the search tree, instead of applying each policy $MaxNbApp$ times alternatively. Concretely, let $Maxdep=min(|V_p|,|V_t|)$. When the tree depth is in range of $[1, \frac{1}{4}Maxdep ]$ and $[\frac{1}{2}Maxdep, \frac{3}{4}Maxdep]$, McSplitDAL+depth uses {\em RL} policy. Otherwise, it uses {\em DAL} policy.

\subsection{Comparison of Performance}
\begin{figure}
	\centering
	\includegraphics[width=3.3in]{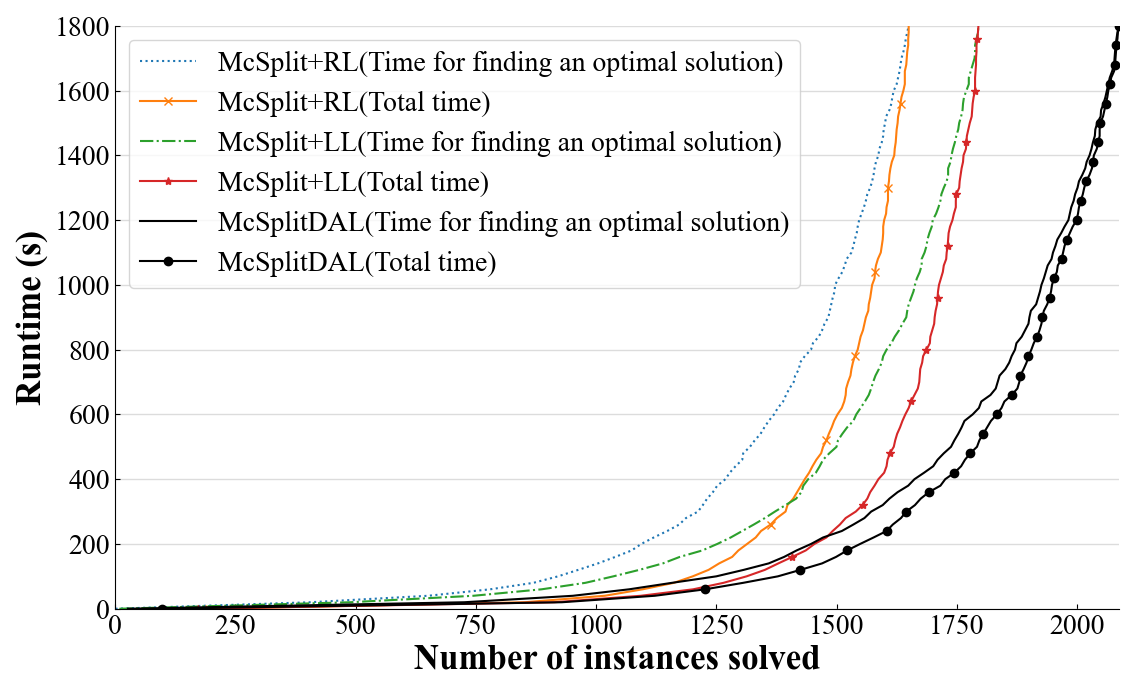}
	\caption {Cactus plots of total instances solved by McSplit+RL, McSplit+LL and McSplitDAL on the 2,229 MCS instances.}%, that solve 1650, 1795 and 2087 instances respectively.}
	\label{Figure2}
\end{figure}

\begin{figure}[t]
	\centering
	\includegraphics[width=3.3in]{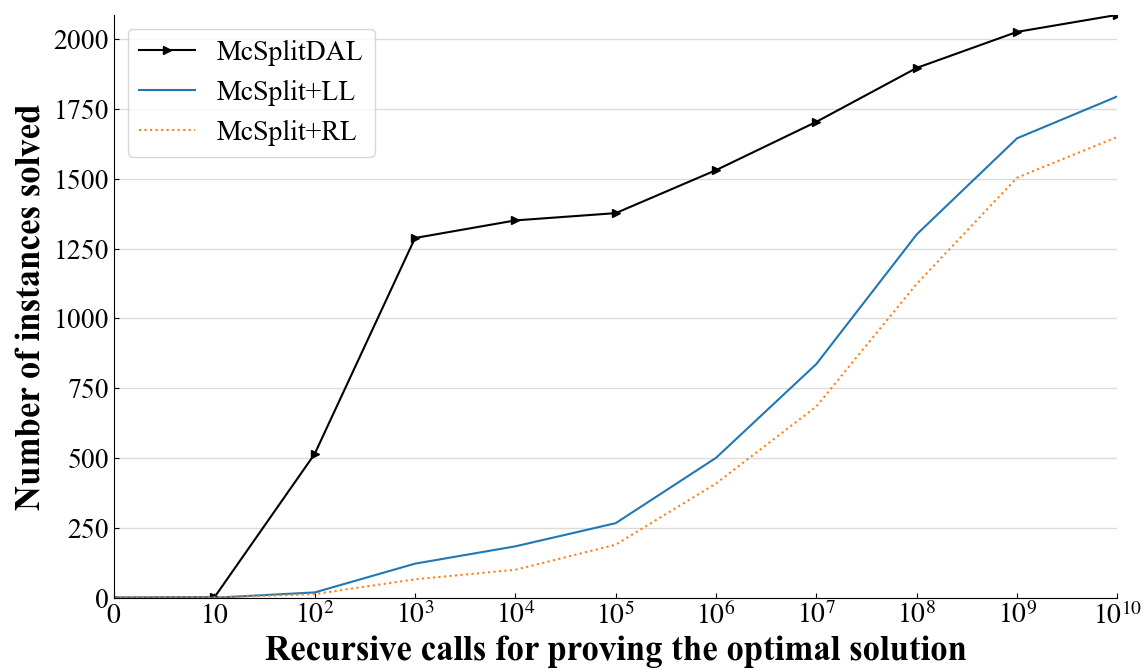}
	\caption {Cactus plots of McSplit+RL, McSplit+LL and McSplitDAL on recursive calls for proving the optimality of the found solution.}
	\label{Figure3}
\end{figure}

\begin{figure}[t]
	\centering
	\includegraphics[width=3.3in]{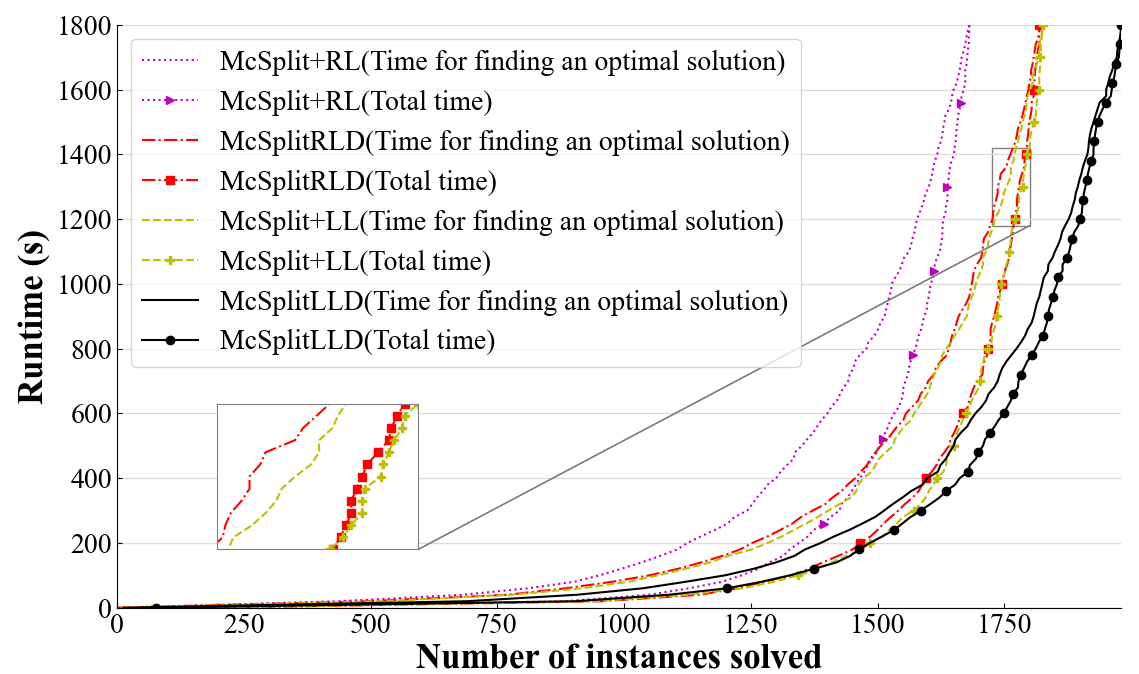}
	\caption {Cactus plots of the total instances solved by McSplit+RL, McSplit+LL, McSplitRLD and McSplitLLD on 2,163 MCS instances.}
	\label{Figure4}
\end{figure}

\begin{figure}[t]
	\centering
	\includegraphics[width=3.3in]{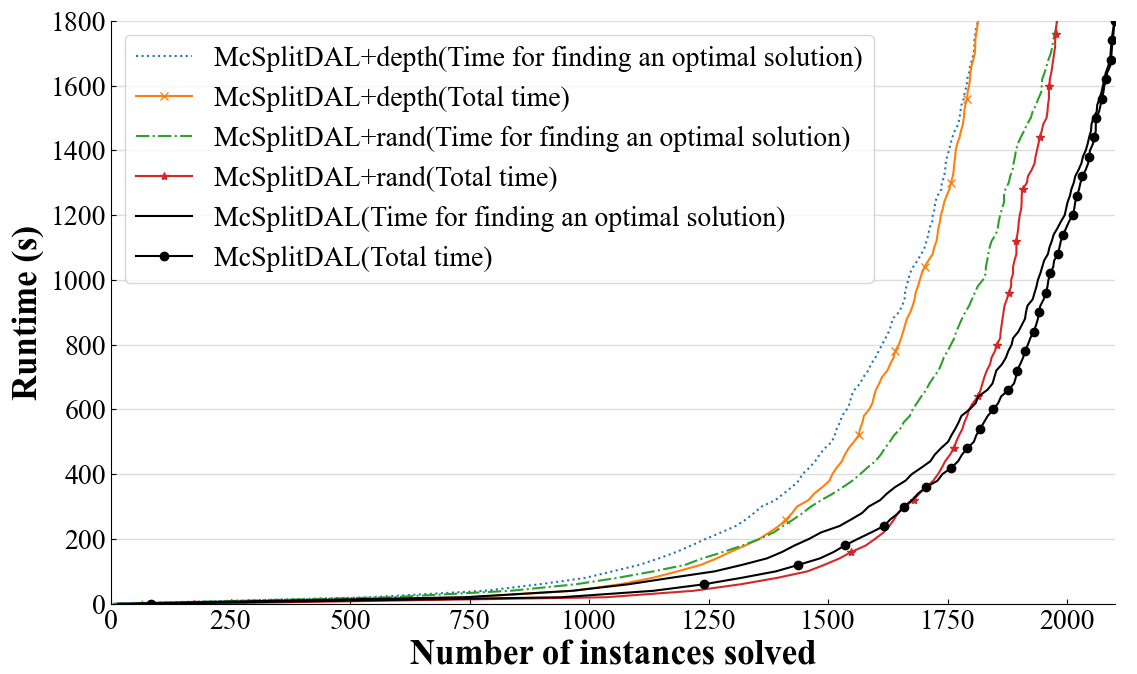}
	\caption {Cactus plots of the number of solved instances by McSplitDAL, McSplitDAL+rand and McSplitDAL+depth on 2,277 MCS instances.}%, that solve 2,100, 1,979 and 1,814 instances respectively.}
	\label{Figure5}
\end{figure}

%Figure \ref{Figure2}
The first experiment compares the general performance of McSplit+RL, McSplit+LL and McSplitDAL on the benchmarks, excluding the too easy instances that can be solved by all the compared solvers within 10 seconds and the too hard instances that cannot be solved by any compared solver within the time limit to make the comparison clearer. %There are 7,716 easy instances solved by the three compared solvers within 10s.
The average runtimes of McSplit+RL, McSplit+LL and McSplitDAL on the excluded easy instances are 0.59s, 0.58s and 0.54s, respectively.

Figure \ref{Figure2} shows the cactus plots of the number of solved instances by the compared three solvers over the remaining 2,229 instances. McSplitDAL solves 292 (437) more instances than McSplit+LL (McSplit+RL). In other words, McSplitDAL solves 16.3$\%$ more instances than McSplit+LL. Note that McSplitDAL, McSplit+LL and McSplit+RL all share the same implementation of Algorithm \ref{alg:BnB} and the unique difference between McSplitDAL and McSplit+LL is the branching heuristic, while the difference between McSplit+LL and McSplit+RL includes the branching heuristic and the LUM technique. However, the performance improvement of McSplitDAL w.r.t. McSplit+LL is greater than the performance improvement of McSplit+LL w.r.t. McSplit+RL. Considering the high performance of baseline algorithms and the NP-hardness of MCS, the results show that new value function {\em DAL} and hybrid branching strategy are very effective for BnB MCS algorithms.

\subsection{Further Analysis}

The search process of an exact MCS algorithm can be divided into two phases: find an optimal solution and prove it is optimal. The experimental results in Figure \ref{Figure2} explains partially why the hybrid learning policy based on the new value function improves the McSplit+LL and McSplit+RL for MCS. For a BnB MCS algorithm, it is easier to reach the pruning condition if the optimal solution is found earlier. Figure \ref{Figure2} shows that McSplitDAL generally finds optimal solutions earlier than McSplit+LL and McSplit+RL, due to the effectiveness of the new value function {\em DAL} and the hybrid vertex selection strategy.

Figure \ref{Figure3} shows the number of recursive calls
%average search tree size
of McSplitDAL, McSplit+LL and McSplit+RL for proving the optimality of the found solution on the same instances as in Figure \ref{Figure2}. The number of recursive calls of McSplitDAL is clearly the smallest, suggesting that the new value function {\em DAL} and the hybrid vertex selection policy in McSplitDAL allows better branching and is also efficient to overcome Matthew effect of a single policy, so that McSplitDAL diversifies better search than McSplit+RL and McSplit+LL, and significantly reduces the number of recursive calls.

\subsection{Ablation Study}

To further access the effectiveness of the proposed
%reward for a matching action (Equation \ref{reward}),
value function, we compare the performance of McSplit+RL with McSplitRLD, and the performance of McSplit+LL with McSplitLLD. The results are showed in Figure \ref{Figure4} (after excluding the too easy instances
%and the too hard instances w.r.t. the four compared algorithms as in Figure \ref{Figure2}).
solved by all the 4 solvers within 10s and the too hard instances that cannot be solved by any of these 4 solvers within 1800s).

Recall that the only difference of McSplit+RL (McSplit+LL) and McSplitRLD (McSplitLLD) is the value function, for McSplitRLD and McSplitLLD do not employ the hybrid vertex selection strategy. As Figure \ref{Figure4} shows, McSplitLLD (McSplitRLD) solves 154 (138) more instances than McSplit+LL (McSplit+RL). In other words, McSplitLLD (McSplitRLD) solves 8.4$\%$ (8.2$\%$) more instances than McSplit+LL (McSplit+RL). So, the results in Figure \ref{Figure4} show that the new value function {\em DAL} is indeed more effective for MCS, because {\em DAL} considers both upper bound reduction and environment changes, while the policies in McSplit+LL and McSplit+RL only consider upper bound reduction.

Note that McSplitDAL solves 138 (299) more instances than McSplitLLD (McSplitRLD), thanks to the hybrid vertex selection strategy (cf. Figure \ref{Figure2}).

The hyper-parameter $MaxNbApp$ in the switching policy conditions is important to McSplitDAL. We leverage McSplitDAL, McSplitDAL+rand and McSplitDAL+depth to evaluate the hybrid policy. Figure \ref{Figure5} shows the comparison of performance of the three solvers over 2,277 instances (after excluding the too easy instances and the too hard instances w.r.t. the three compared algorithms as before). The comparison shows that McSplitDAL has the best performance, solving 121 and 286 more instances than McSplitDAL+rand and McSplitDAL+depth, respectively. The goal of MCS algorithm is to find an optimal solution which size is at most $min(|V_p|,|V_t|)$. Experimental results show the switch condition related to the optimal solution size have a better performance than random choice and fixed tree depth.

\section{Conclusion}
In this paper, we propose a new value function and a hybrid branching strategy in a branch-and-bound (BnB) algorithm based on reinforcement learning for the Maximum Common induced Subgraph (MCS) problem. The new value function considers both upper bound reduction and environment change to reward an action of matching two vertices. It allows to select vertices to better simplify the graphs. The hybrid branching strategy guides the search by employing alternatively two different branching heuristics to diversify the search and find optimal solutions earlier. We implement the new approaches into a BnB algorithm called McSplitDAL. Extensive experimental results show that the proposed methods significantly improve the efficiency of the BnB MCS algorithm, and McSplitDAL solves the highest number of instances.

In the future, we will continue to study the interplay of the search and the learning, and apply our approach to solve other graph matching problems.
% Use \bibliography{yourbibfile} instead or the References section will not appear in your paper
%\nobibliography{aaai23}
\bibliography{ms}
%\fontsize{9.0pt}{10.0pt} \selectfont

\end{document}